\documentclass[11pt,a4paper]{article}
\usepackage[hyperref]{emnlp2020}
\usepackage{times}
\usepackage{xcolor}
\usepackage{latexsym}
\usepackage{hyperref}
\usepackage{amsfonts}
\usepackage{amsmath} 
\usepackage{graphicx}
\usepackage{subcaption}
\usepackage{bbm}
\usepackage{multirow}
\usepackage{booktabs,arydshln}
\usepackage{float}
\usepackage{comment}

\makeatletter
\def\adl@drawiv#1#2#3{%
        \hskip.5\tabcolsep
        \xleaders#3{#2.5\@tempdimb #1{1}#2.5\@tempdimb}%
                #2\z@ plus1fil minus1fil\relax
        \hskip.5\tabcolsep}
\newcommand{\cdashlinelr}[1]{%
  \noalign{\vskip\aboverulesep
           \global\let\@dashdrawstore\adl@draw
           \global\let\adl@draw\adl@drawiv}
  \cdashline{#1}
  \noalign{\global\let\adl@draw\@dashdrawstore
           \vskip\belowrulesep}}
\makeatother

\usepackage{microtype}

\aclfinalcopy % Uncomment this line for the final submission
\newcommand{\note}[1]{\textcolor{red}{(#1)}}

\newcommand{\cmmnt}[1]{\ignorespaces}

\usepackage{color, colortbl}
\definecolor{LightCyan}{rgb}{0.88,1,1}

\usepackage{lipsum}

\title{Guiding Attention for Self-Supervised Learning with Transformers}

\author{Ameet Deshpande\quad Karthik Narasimhan \\
  Department of Computer Science \\
  Princeton University \\
  \texttt{\{asd, karthikn\}@cs.princeton.edu}}
  
\date{}

\begin{document}
\maketitle
\begin{abstract}
% Despite being successful in downstream language understanding tasks, modern language models~\cite{gpt, bert, roberta} contain millions of parameters and require multiple days of training on specialized hardware such as TPUs. Training such models on commodity hardware (e.g. GPUs) often means slow convergence, making it practically intractable for many researchers. 
In this paper, we propose a simple and effective technique to allow for efficient self-supervised learning with bi-directional Transformers. Our approach is motivated by recent studies demonstrating that self-attention patterns in trained models contain a majority of non-linguistic regularities. We propose a computationally efficient auxiliary loss function to guide attention heads to conform to such patterns. Our method is agnostic to the actual pre-training objective and results in faster convergence of models as well as better performance on downstream tasks compared to the baselines, achieving state of the art results in low-resource settings. Surprisingly, we also find that linguistic properties of attention heads are not necessarily correlated with language modeling performance.\footnote{Code: \href{https://github.com/ameet-1997/AttentionGuidance}{https://github.com/ameet-1997/AttentionGuidance}}
\end{abstract}

% \section*{Outline}
% \begin{itemize}
%     \item Current pre-training methods are too expensive, require days of training with multiple TPUs. 
%     \item Recent studies have investigated the various attention patterns learnt by models like BERT, GPT, etc. and surprisingly, most of the attention heads seem to learn non-linguistic patterns and yet seem to help downstream task performance after fine-tuning. (Need to cite all the work that helps make these points).
%     \item In fact, even if one takes a pre-trained model from a different language (e.g. French/Chinese) and fine-tunes on English GLUE tasks, the performance is quite good. 
%     \item This brings us to the key question: What attention patterns actually help initialize the Transformer-based models well enough to provide downstream gains? Can we initialize these attention patterns mathematically (or through auxiliary losses) to reduce the need for long and expensive computation regimes?
%     \item Further, would our approach help in low-resource domains where there is not enough data to pre-train in the first place?
%     \item Experiments and results to investigate the above questions.
% \end{itemize}

\section{Introduction}
\label{sec:introduction}

Recent advances in self-supervised pre-training~\cite{radford2018improving, devlin2018bert, liu2019roberta} have resulted in impressive downstream performance on several NLP tasks~\cite{wang2018glue,wang2019superglue}. However, this has led to the development of enormous models, which often require days of training on non-commodity hardware (e.g. TPUs)~\cite{kaplan2020scaling}. Furthermore, studies have shown that it is quite challenging to successfully train these large Transformer models~\cite{vaswani2017attention}, requiring complicated learning schemes and extensive hyperparameter tuning~\cite{xiong2020layer,raffel2019exploring,popel2018training}.

Despite these expensive training regimes, recent studies have found that once trained, these bi-directional language models exhibit simple patterns of self-attention without much linguistic backing~\cite{voita2019analyzing,raganato2018analysis}. For example, 40\% of heads in a pre-trained BERT model~\cite{devlin2018bert} simply pay attention to delimiters added by the tokenizer (e.g. \texttt{[CLS]} or \texttt{[SEP]})~\cite{kovaleva2019revealing}. Since these attention patterns are independent of linguistic phenomena, a natural question arises: can Transformer models be guided towards such attention patterns without requiring extensive training?

\begin{figure}[t]
% \vskip 0.2in
\begin{center}
\centerline{\includegraphics[width=\linewidth]{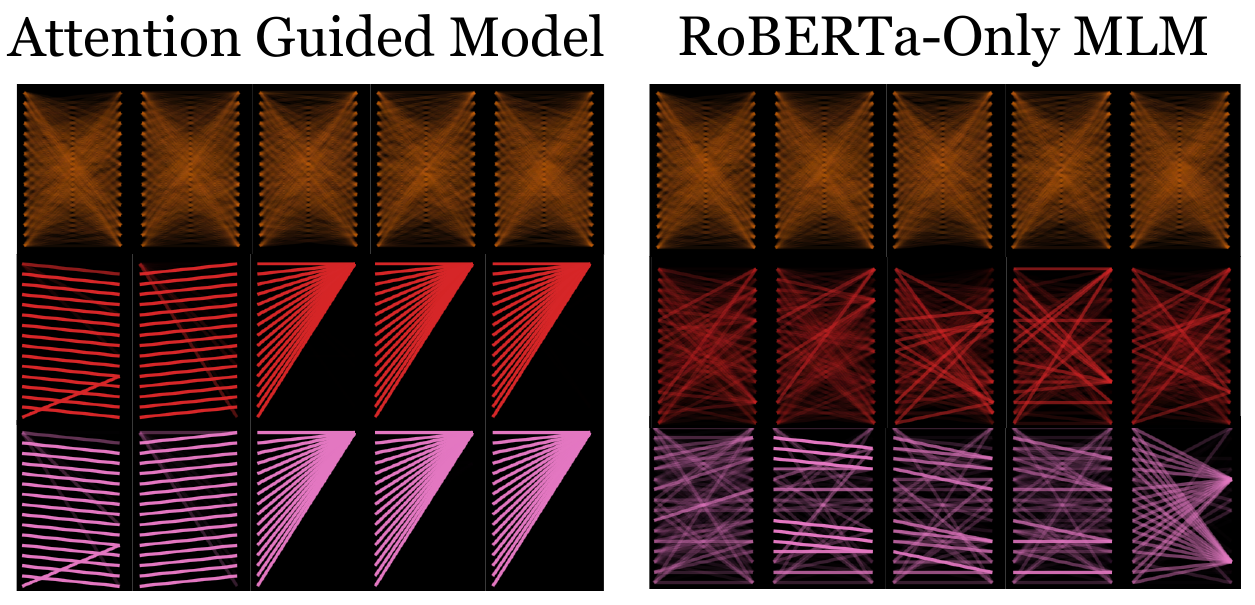}}
\caption{Attention patterns of our model (left) and the default RoBERTa model (right) after 0\% (top), 1\% (middle) and 100\% (bottom) of pre-training. Inducing simple patterns (left) using an auxiliary loss leads to benefits in convergence speed, downstream performance, and robustness to hyperparameters.}
\label{fig:attentionpatterns1}
\end{center}
\vskip -0.4in
\end{figure}

In this paper, we propose an \textit{attention guidance} (AG) mechanism for self-attention modules in Transformer architectures to enable faster, more efficient, and robust self-supervised learning. Our approach is simple and agnostic to the training objective. Specifically, we introduce an auxiliary loss function to guide the self-attention heads in each layer towards a set of pre-determined patterns (e.g. Figure~\ref{fig:attentionpatterns1}~\cite{vig2019multiscale}). These patterns encourage the formation of both  global (e.g. attend to \texttt{[CLS]}, \texttt{[SEP]} tokens) and local (e.g. attend to \texttt{[Next]}, \texttt{[Prev]} token) structures in the model.

Through several experiments, we show that our approach enables training large Transformer models considerably faster — for example, we can train a 16-layer RoBERTa model with SOTA performance on a low-resource domain in just two days using four GPUs, while excluding our loss leads to slow or no convergence. Our method also achieves competitive performance with BERT~\cite{devlin2018bert} on three English natural language understanding tasks, and outperforms the baseline \textit{masked language modeling} (MLM) models on eleven out of twelve settings considered.

Further, we also show that our initialization is agnostic to the training objective by demonstrating gains on the \textit{replaced token detection} objective proposed by ELECTRA~\cite{clark2020electra} and on machine translation with Transformers. Finally, we provide an analysis of the attention heads learned using our method. Surprisingly, contrary to recent studies~\cite{clark2019does, lin2019open}, we find that it is possible to train models that perform well on language modeling without learning a single attention head that models coreferences.
% . For example, our model fails the co-reference test in \cite{clark2019does} while still performing well on language modeling and downstream tasks.

To summarize, our main contributions are:
\begin{itemize}
	\item We propose a simple auxiliary loss for self-attention heads that enables large models to converge quickly on commodity hardware.
    \item We demonstrate the effectiveness of our auxiliary loss on different languages, model sizes, and training objectives.
    \item We provide evidence that the linguistic performance of individual attention heads is not a necessary condition for good language modeling (LM) or downstream task performance.
\end{itemize}
\section{Related Work} \label{sec:related_work}

%\paragraph{Transformers for language modeling}
%Since OpenAI GPT \cite{radford2018improving} and BERT \cite{devlin2018bert} popularized the Transformer~\cite{vaswani2017attention} for self-supervised learning from raw text, several studies have critically evaluated each component of these models and proposed variants like GPT-2 \cite{radford2019language}, RoBERTa~\cite{liu2019roberta}, XLNet~\cite{}, and T5~\cite{raffel2019exploring}. In other work, \citet{sun2019ernie} proposed a lifelong multi-task learning setup (ERNIE 2.0) with a focus on language-understanding on downstream tasks. \citet{lan2019albert}  proposed ALBERT -- a variant of BERT with weights shared across all layers. 

\paragraph{Improving efficiency of LMs}
The high computational costs of BERT-style models have accelerated research on developing efficient contextual language models. \citet{clark2020electra} used a GAN-like setup to predict if each word in the input sequence is corrupted by a generator (another pre-trained LM). They show that their method is more sample efficient than the standard MLM objective. Other studies have explicitly focused on making the self-attention modules more efficient. Reformer~\cite{kitaev2020reformer} and Sparse Transformer~\cite{child2019generating} introduce locality-sensitive hashing and sparse factorizations to reduce the quadratic complexity of dot-product attention, while Longformer~\cite{beltagy2020longformer} uses local-windowed and task motivated global attention to scale the memory usage of self-attention modules linearly.
% TinyBERT~\cite{jiao2019tinybert} uses a two-stage distillation framework to learn from a BERT ``teacher" model. DistilBERT~\cite{sanh2019distilbert} uses knowledge distillation during pre-training and retains a large fraction of BERT's performance.

\paragraph{Analyzing Self-Attention}
Recent papers have analyzed the attention patterns in trained Transformer-based LMs. Some studies hypothesize that multiple attention heads capture linguistic phenomena like co-reference links and dependency arcs~\cite{clark2019does,htut2019attention}. However, other studies show that pruning those heads leads to minimal performance degradation on downstream tasks~\cite{kovaleva2019revealing,michel2019sixteen}. Others note that there are recurring patterns in attention distributions corresponding to different attention heads (hereon, heads), which are not language or task-dependent~\cite{voita2019analyzing,raganato2018analysis}. While our study also questions the role of heads for language modeling and downstream performance, we focus on making modifications to the LM pre-training and not on analyzing published pre-trained models.

% This points to 2 things -- 1) there is a lot of redundancy in the attention patterns in Transformer-based LMs, and 2) the linguistic capabilities of self-attention heads is not a given even if the LM achieves very good perplexity scores.    

% a pre-trained transformer's self-attention \cite{vaswani2017attention}. \cite{clark2019does}, \cite{kovaleva2019revealing}, \cite{voita2019analyzing}, and \cite{raganato2018analysis} note that there are recurring patterns in attention distributions corresponding to different attention heads (hereon, heads) which are not language or task dependent. While \cite{clark2019does} and \cite{htut2019attention} hypothesize and show that there are multiple attention heads which could be performing linguistically motivated tasks like co-reference resolution and subsets of dependency parsing, \cite{kovaleva2019revealing} note that pruning those heads leads to no performance degradation on downstream tasks. \cite{michel2019sixteen} use greedy algorithms to prune a large number of heads while suffering very little deterioration. This redundancy in attention is exploited in BlockBERT \cite{qiu2019blockwise} and Longformer \cite{beltagy2020longformer} where attention is paid mainly to a local context only. However, these implementations need significant changes to the model.

\paragraph{Constraining Self-Attention}
\citet{qiu2019blockwise} enforce local constraints on the attention patterns to reduce computation and build deeper models with longer contexts.
% ~\cite{child2019generating,qiu2019blockwise,beltagy2020longformer}.
The studies that are perhaps most similar to ours explore fixed attention patterns for machine translation~\cite{you2020hard,raganato2020fixed}. \citet{you2020hard} replace all attention heads in the encoder with hard-coded Gaussian distributions centered around the position of each token while observing a minimal reduction in BLEU scores. \citet{raganato2020fixed} substitute all but one head with fixed attention patterns in each encoder layer and note little performance degradation. Both these studies enforce hard constraints on the self-attention and try to match baselines in terms of speed and performance. Our approach is complementary -- our attention guidance loss is a form of soft regularization and outperforms baseline models both in terms of convergence speed and quantitative metrics.

% \paragraph{Papers analyzing BERT attention}

% \paragraph{BERT in different languages}
% - Have similar attention patterns (Chinese, French, German, etc.)
% - Can be finetuned even on English tasks

% \paragraph{Case for Hard attention}
% - REFORMER
% - Longformer
% - BlockBERT

% \paragraph{Other Attention Analysis Studies}
% Make the case that attention need not correspond to respective tokens in any layer other than the first layer. Either cite other papers, or run experiments to show this. Also cite Clark's paper and ``Dark Secrets" paper

% \paragraph{Evidence for hypothesis}
% \begin{itemize}
%     \item Longformer: The Long-Document Transformer
% \end{itemize}

% \paragraph{Redundancy in BERT}

% \begin{enumerate}
%     \item \href{https://arxiv.org/pdf/2004.04010.pdf}{Exploiting Redundancy in Pre-trained Language Models for Efficient Transfer Learning}
%     \item ALBERT paper
% \end{enumerate}
\section{Approach}
\label{sec:approach}

\subsection{Prelude: The surprising effectiveness of non-linguistic attention}\label{sec:motivation}

Several recent studies~\cite{clark2019does,kovaleva2019revealing} have demonstrated that Transformers trained with the masked language modeling (MLM) objective exhibit simple self-attention patterns (e.g., attending to delimiter tokens). These patterns (e.g. Figure \ref{fig:attentionpatterns}) are consistent across models pre-trained on different languages, or fine-tuned on various downstream tasks~\cite{kovaleva2019revealing}. Since these patterns are not linguistically motivated, we hypothesize that pre-training a model serves the dual purpose of lending linguistic and non-linguistic structure. To test the impact of the latter, we finetune CamemBERT~\cite{martin2019camembert} (a model pre-trained on the French part of OSCAR corpus~\cite{suarez2019asynchronous}), and BERT-Base Chinese~\cite{bertbasechinese} (a model pre-trained on Chinese Wikipedia articles), on three English downstream tasks~\cite{socher2013recursive,rajpurkar2016squad,dolan2005automatically}. We also compare with a randomly initialized Transformer, which is finetuned on downstream tasks without any pre-training \cite{kovaleva2019revealing}.

Surprisingly, the results in Table \ref{table:difflanguage} show that despite both models having mismatched tokens and being trained on languages with linguistic constructs that are different from those of English, their performance is significantly better than a model with no pre-training. This corroborates the idea that the non-linguistic structure in attention heads is beneficial for learning, and inducing it explicitly may lead to faster training and better performance.

\begin{table}
\centering
\resizebox{1\columnwidth}{!}{
\begin{tabular}{c c c c}
\toprule
\textbf{Pre-trained (PT) Model} & \textbf{SST-2} & \textbf{MRPC} & \textbf{QNLI} \\ \midrule 
No PT \cite{kovaleva2019revealing} & 0.80 & 0.81/0.68 & 0.49 \\ \cdashlinelr{1-4}
Chinese PT \cite{bertbasechinese} & 0.86 & 0.86/0.81 & 0.83  \\
% RoBERTa-Attn ($\lambda=1/2$) & 0.674 &  0.047 & 0.704 & 0.047 \\
French PT \cite{martin2019camembert} & 0.88 & 0.88/0.84 & 0.85 \\ \cdashlinelr{1-4}
English PT \cite{devlin2018bert} & 0.93 & 0.89/0.84 & 0.91 \\ \bottomrule
\end{tabular}
}
\caption{Models pre-trained even with French and Chinese data perform significantly better than no pre-training on \textit{English} downstream tasks.}
\label{table:difflanguage}
\end{table}

\subsection{Our method: Attention guidance for Transformers} \label{section:method}
We first formally define the masked language modeling (MLM) setup with Transformers~\cite{vaswani2017attention} and then describe our attention guidance mechanism.

\paragraph{MLM with Transformers}
Transformers used for sequence-to-sequence prediction tasks are trained on a dataset $\mathcal{D}$ of pairs of sequences $\mathbf{x}$ and corresponding labels $\mathbf{y}$. In the case of masked language modeling (MLM), the input sequence $x_1, x_2, \dots ,x_n$ of length $n$ consists of individual tokens and the output labels $y_1, y_2, \dots, y_n$ are the same as the input sequence, i.e.,  $y_i = x_i$. A fraction $k$ of the input tokens, chosen randomly, are \emph{masked}, i.e., replaced with a \texttt{$<$MASK$>$} token. Assume that these masked indices are collected together in a set $\mathcal{C}$. The MLM objective then is a cross-entropy loss on the predictions $y'_j$ made by the model at the masked locations $j\in \mathcal{C}$, and is used to optimize all the parameters of the model, $\theta$ by minimizing:
$$\mathcal{L}_{MLM}(\mathbf{x}, \mathbf{y}) = - \sum_{j\in \mathcal{C}} \log \mathbb{P}\left (y_j|\mathbf{x};\theta \right ) $$

The Transformer architecture for MLM consists of $\ell$ layers with $h$ self-attention heads per layer. Let the input activations to layer $k$ of this model be $\textbf{s}_k$, with $|\textbf{s}_k|=n$. Naturally, $\textbf{s}_1=\textbf{s}=\textbf{x}$. For every position $p \in [1,n]$ in its output, each attention head in layer $k$ induces a probability distribution over all positions in the input $\textbf{s}_k$. Let a single head's attention activations (as described in Equation $1$ of \cite{vaswani2017attention}), which is a function of $\mathbf{s}$, be denoted by the following:

\begin{equation}
\mathbf{H}(\mathbf{s})  = \textrm{softmax}\left (\dfrac{QK^{\top}}{\sqrt{d_k}}\right ) \in \mathbb{R}^{n\times n},
  \label{eq:head_activations}    
\end{equation}
where $Q$ and $K$ are the query and key matrices respectively, and $d_k$ is the dimensionality of the queries or keys.
Further, let $\mathbf{H}(\mathbf{s})[p,q]$ (a scalar) denote the attention that token $p$ in the head's output layer pays to token $q$ in the head's input layer. We drop the dependence on $\mathbf{s}$ in the following sections for notational convenience.

\paragraph{Guiding attention heads}

%Requirements:
%1. don't change the objective, agnostic
%2. manipulate attention patterns
%3. faster training, better performance

%Given the redundancy in the attention heads noted in section \ref{sec:related_work}, we guide a fraction of them in each layer to conform to a specific pattern. \cite{clark2019does} and \cite{kovaleva2019revealing} note that a large number of heads pay attention mainly to delimiter tokens like \texttt{[CLS]}, \texttt{[SEP]}, and \texttt{[Period]} (`.') \note{make these model agnostic} which are either added by models, or are frequent in a language. Further, words in a fixed context window are crucial to predict a masked word in MLM. This can be seen in the sentence, ``\textit{The} $<$\texttt{MASK}$>$ \textit{barks at the cat}", where the word \textit{barks} enables the model to predict the label \textit{dog} without much other context. 

To guide an attention head, we impose a mean squared error (MSE) loss on $\mathbf{H}$ using a pre-defined pattern $\mathbf{P}(\mathbf{s}) \equiv \mathbf{P} \in \mathbb{R}^{n\times n}$, where $||\cdot||_\text{F}$ is the Frobenius norm:
\begin{equation}
  \mathcal{L}_{attn}\left (\mathbf{H}, \mathbf{P}\right ) = ||\mathbf{H}-\mathbf{P}||_\text{F}^2
  \label{eq:attn-loss}    
\end{equation}
Specifically, we consider two types of patterns:
\begin{itemize}
    \item \textbf{Global attention patterns} that focus their attention on specific global positions like the first token of the sequence (\texttt{[First]}), punctuations like the \textit{period} token (\texttt{[Period]}), or on delimiters like \texttt{[CLS]} or \texttt{[SEP]} added by the tokenizer (\texttt{[Delim]}). As an example,
$\mathbf{P}_{\textrm{\texttt{[First]}}}[p,q] = \begin{cases} 1 &\mbox{if } q = 1 \\ 
0 & \textrm{otherwise} \end{cases}$ 
    
    % $\mathbf{P}_{\textrm{\texttt{[First]}}}[p,q]=1$ if $q=1$, else $0$
    \item \textbf{Local attention patterns} that focus \textit{either} on the next or the previous token (e.g. \texttt{[Next]}, \texttt{[Prev]}). As an example,  
    
    $\mathbf{P}_{\textrm{\texttt{[Next]}}}[p,q] = \begin{cases} 1 &\mbox{if } q = p+1 \\
    \frac{1}{n} &\mbox{if } p = n \\
    0 & \textrm{otherwise} \end{cases}$
\end{itemize}

\begin{figure*}[ht]
% \vskip 0.2in
\begin{center}
\centerline{\includegraphics[width=\linewidth]{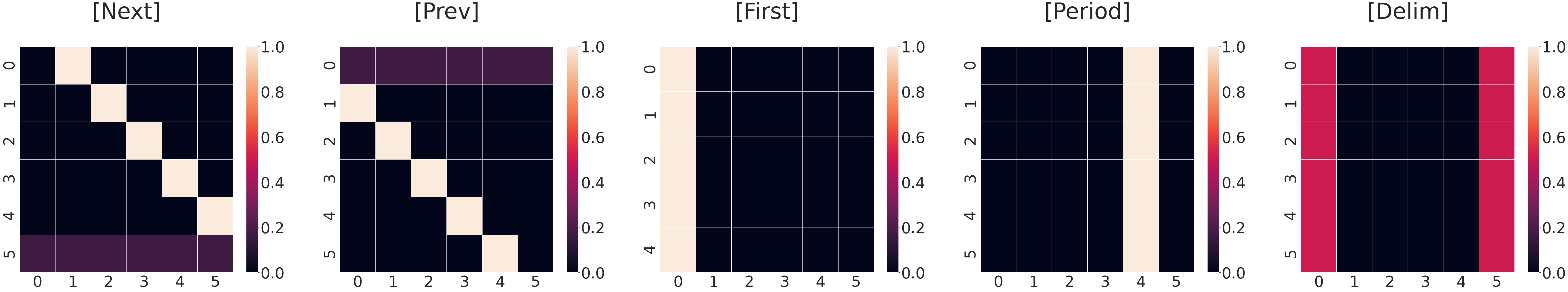}}
\caption{Example attention patterns used in our AG models for the sentence ``\textit{$<$s$>$ Welcome to EMNLP . $</$s$>$}". Note that the first three patterns (\texttt{[Next]}, \texttt{[Prev]}, \texttt{[First]}) do not even depend on the input sentence.}
\label{fig:attentionpatterns}
\end{center}
\vskip -0.2in
\end{figure*}

Figure \ref{fig:attentionpatterns} displays example $\mathbf{P}$ matrices for the different patterns we use. Note that the first and the last rows in \texttt{[Prev]} and \texttt{[Next]} patterns respectively are set to uniform distributions.

% While all the desired patterns are mathematically simple to express, we add an auxiliary loss because the varying distribution of inputs caused the change in parameters makes it difficult to guide them to specific patterns without any training. This also allows the procedure to be model agnostic. To satisfy these desiderata, we propose to use auxiliary losses on the intermediate attention layers to guide them towards desired patterns.

\paragraph{Overall loss function}
We apply the attention loss in Equation~\ref{eq:attn-loss} to each head in each layer to obtain the overall \textbf{attention guidance (AG) loss}:
\begin{equation}
    \mathcal{L}_{AG} (\mathbf{x}) = \sum_{k=1}^\ell \sum_{j=1}^h  \mathcal{L}_{attn}\left (\mathbf{H}_{kj}, \mathbf{P}_{kj} \right ) \times \mathbbm{1}(k, j),
\end{equation}
where $\mathbbm{1}(k,j)$ denotes an indicator function which is $1$ only if the $j$th head in layer $k$ is being guided. 

In general, this loss allows for arbitrary choices of patterns for each $\mathbf{P}_{kj}$. However, to simplify matters in our experiments, we guide a particular head number to the same pattern across all layers, i.e., $\mathbf{P}_{\cdot j}$ is the same for all layers. We utilize the gradients from this loss to update all the parameters of the model (including the feedforward and input embedding layers). It is worth noting that this loss only depends on the input $\mathbf{x}$ and not on labels $\mathbf{y}$.

Finally, we combine our attention guidance (AG) loss with the MLM loss to get our overall optimization objective:
\begin{equation} \label{equation:alpha}
    \mathcal{L}(\theta) = \mathbb{E}_{(\mathbf{x}, \mathbf{y}) \sim \mathcal{D}}\left [\mathcal{L}_{MLM}(\mathbf{x},\mathbf{y}) + \alpha_t \cdot \mathcal{L}_{AG}(\mathbf{x})\right ]
\end{equation}
where $\alpha_t$ is a hyperparameter. In practice, we find that $\mathcal{L}_{AG}$ converges faster than $\mathcal{L}_{MLM}$, so we linearly decay $\alpha_t$ from an initial value $\alpha_0$ to $0$ as the training progresses (details in Section~\ref{sec:experiments}).

%Note that we add the attention loss as an auxiliary loss instead of enforcing hard attention so that the attention heads can violate the patterns if it is useful for reducing $\mathcal{L}_{MLM}$.

\section{Experimental Setup}
\label{sec:experiments}

We demonstrate the effectiveness of our attention guidance loss through several empirical studies. Specifically, we 1) report convergence results on masked language modeling, 2) evaluate trained language models on downstream tasks, and 3) analyze the learned attention representations using probes. For 1) and 2) above, we perform experiments on both high-resource and low-resource settings. 

\subsection{Datasets}

%To show the usefulness of our method, we carry out experiments on datasets of different sizes. The choice of the datasets is motivated by the presence of downstream tasks.
We use the following datasets spanning three different languages (details in Table~\ref{table:dataset_stats_1}):
\begin{enumerate}
	\item \emph{English}: To train language models, we use a 2.1 billion token corpus from English Wikipedia. We download and pre-process articles according to \citet{shoeybi2019megatron}. For downstream evaluation, we choose  three tasks: QQP\footnote{\url{https://www.quora.com/q/quoradata/First-Quora-Dataset-Release-Question-Pairs}}, MNLI~\cite{williams2017broad}, and QNLI~\cite{rajpurkar2016squad}
	\item \emph{Filipino}: We use a 36 million token corpus of Wikipedia text collected by~\citet{cruz2020establishing} to train language models, and the accompanying binary sentiment classification task to evaluate downstream performance.
	\item \emph{Oromo}: Our smallest corpus contains 4.6 million tokens (\cite{strassel2016lorelei}). We use the accompanying named entity tags for NER, which is our downstream task.
\end{enumerate}
These cover a range of dataset sizes — from high-resource (English) to low-resource (Oromo). 

\begin{table}
\centering
\resizebox{1\columnwidth}{!}{
\begin{tabular}{c c c c c c}
\toprule
\multirow{2}{*}{\textbf{Lang.}} & \multicolumn{2}{c }{\textbf{LM training}} & \multicolumn{3}{c }{\textbf{Downstream task}} \\ \cmidrule(lr){2-3} \cmidrule(lr){4-6}  
 & \textbf{Train} & \textbf{Valid} & \textbf{Task} & \textbf{\# instances} & \textbf{Eval. Metric} \\ \midrule

\multirow{3}{*}{English} & \multirow{3}{*}{2116M} & \multirow{3}{*}{1\%} & MNLI & 393k & Accuracy\\
&  & & QNLI & 105k & Accuracy\\
&  & & QQP & 364k & F-1\\ \cdashlinelr{1-6}
Filipino & 36M & 10\% & SC & 10k & Accuracy\\
Oromo & 4.6M & 3\% & NER & 1k & F-1\\ \bottomrule
\end{tabular}
}
\caption{Dataset statistics for LM training and downstream tasks on English, Filipino, and Oromo. SC=Sentiment Classification. Filipino and Oromo are low-resource languages.}
\label{table:dataset_stats_1}
\end{table}

\subsection{Evaluation}
%\note{Emphasize that we report MLM, even though we are training with an auxiliary loss}
Evaluation metrics for the different tasks:
\begin{enumerate}
	\item \emph{Language modeling}: We report the training and validation MLM losses. Even though our attention guided models are trained with an auxiliary loss, we report only the MLM loss for direct comparison with the corresponding baseline. We also report the average training loss to compare models' convergence rates.
	\item \emph{English downstream tasks}: Accuracy for MNLI and QNLI, and F-1 score for QQP.\footnote{We report the test scores obtained by submitting to \url{https://gluebenchmark.com}~\cite{wang2018glue}}
	\item \emph{Filipino Sentiment Classification}: Since the dataset for Filipino is balanced, we use binary classification accuracy.
	\item \emph{Oromo NER}: Following \citet{wang2020extending}, we perform 10-fold cross-validation and use the F-1 scores aggregated over 9 tag classes.
% 	\item \emph{Downstream tasks}: We use standard evaluation metrics provided in GLUE for English, accuracy for sentiment classification in Filipino, and F-1 scores for NER in Oromo
\end{enumerate}

\subsection{Models and Training}\label{sec:model_definitions}

% \karthik{I rephrased things here a bit. double check this paragraph to make sure details are correct and remove this comment.}
To make comparisons across different settings easy, we choose RoBERTa~\cite{liu2019roberta} as the base architecture for all our experiments. We train variants with $8,12$, and $16$ layers following the configurations given in the original paper~\cite{liu2019roberta} on all 3 languages, which gives us a total of 9 settings. Since the current SOTA model for Filipino~\cite{cruz2019evaluating} is a BERT model, we train our Filipino models with both the MLM and next sentence prediction loss. Details of the model hyperparameters are provided in Appendix~\ref{appendix:hyperparameters}. For each model, we compare its learning with and without our AG loss. We denote the attention guided models by RoBERTa-AG and the unmodified versions by RoBERTa-MLM. For notational convenience, RoBERTa-$X$-MLM and RoBERTa-$X$-AG represent RoBERTa models with $X$ layers. 

\paragraph{Comparison with state-of-the-art (SOTA)} While we train all variants of our models with and without AG loss, and only these results are strictly comparable, we also compare with SOTA models for reference. These are E-MBERT~\cite{wang2020extending}, a recent extension of multilingual BERT~\cite{devlin2018multi} which performs well on low-resource languages, for Oromo, BERT~\cite{cruz2020establishing} for Filipino, and RoBERTa$_{\textrm{BASE}}$~\cite{liu2019roberta} for English\footnote{MNLI-m and MNLI-mm scores are reported as the same in table~\ref{table:downstream} because they are not reported separately in~\cite{liu2019roberta}. QQP scores reported are for RoBERTa$_\textrm{Large}$ because the F-1 score is not reported for RoBERTa$_\textrm{Base}$}.

\subsection{Attention patterns}
We consider the following patterns $\mathbf{P}$ (section \ref{sec:approach}) for guiding the self-attention heads:
\begin{enumerate}
    \item \texttt{[Next]} attends to the next token.
    \item \texttt{[Prev]} attends to the previous token.
    \item \texttt{[First]} attends to the first token in the sequence.
    \item \texttt{[Delim]} attends to delimiter tokens like \texttt{<s>}, \texttt{</s>}, \texttt{[CLS]} and \texttt{[SEP]} added by the model's tokenizer.
    \item \texttt{[Period]} attends to the period (`.') token.
\end{enumerate}

Only the \texttt{[Delim]} and \texttt{[Period]} patterns depend on the input because the corresponding tokens vary in position with the input. All other patterns are static and have a low memory footprint. Mathematical specifications of these patterns are provided in appendix \ref{appendix:mathspec}, and Figure~\ref{fig:attentionpatterns} illustrates them.

\subsection{Implementation details}\label{section:implementation}
\textbf{Basic MLM models:}\quad
We tune the learning rate from the set $\{$1e-5, 5e-5, 1e-4$\}$, the dropout in self-attention from the set $\{0.0, 0.1\}$, and the number of warmup steps from the set $\{0, 1000, 10000\}$.\\
\textbf{AG models:}\quad For our AG models, we guide a fraction $\lambda \in \{\frac{1}{4}, \frac{2}{4}, \frac{3}{4}, 1 \}$ of heads in each layer. We choose $\alpha_0$ (equation \ref{equation:alpha}) from the set $\{1, 10, 100 \}$ such that the scales of the MLM loss and auxiliary loss are comparable at the beginning of training. 

\paragraph{Best performing hyperparameters:}
\textbf{RoBERTa-MLM} is very sensitive to the learning rate and the number of warmup steps, and the best performing hyperparameters are reported in appendix \ref{appendix:hyperparameters}. On the other hand, we find that \textbf{RoBERTa-AG} is very robust and does not need much tuning. A learning rate of 1e-4, $\lambda=0.5$, and $0$ warmup steps work well for all the experiments. $\alpha=10$ is used for our 12,16 layer models, and $\alpha=100$ for smaller models. We fit the largest batch size possible for each model. We perform an ablation study and find that the \texttt{[Next]} and \texttt{[Prev]} patterns are most important, followed by \texttt{[First]}(section~\ref{appendix:pattern_ablation}). Hence, one head each is modified with the \texttt{[Next]} and \texttt{[Prev]} patterns, and $(\lambda h-2)$ heads are modified with \texttt{[First]}.
% \textbf{RoBERTa-MLM} is very sensitive to the learning rate and the number of warmup steps, and the best performing hyperparameters are reported in appendix \ref{appendix:hyperparameters}. On the other hand, we find that \textbf{RoBERTa-AG} is very robust and does not need much tuning. A learning rate of 1e-4, $\lambda=\frac{1}{2}$, and $0$ warmup steps work well for all the experiments. $\alpha=10$ is used for our 12,16 layer models, and $\alpha=100$ for smaller models. One head each is modified with the \texttt{[Next]} and \texttt{[Prev]} patterns, and $(\lambda h-2)$ heads are modified with \texttt{[First]}. We observed that adding \texttt{[Delim]} led to no significant change in performance and \texttt{[Period]} led to a small decrease.

\paragraph{Compute Time and Hardware} Unlike state-of-the-art models, we emphasize that our studies are performed on a smaller computational budget, both with respect to wall clock time and hardware. Our English models are trained for 10 epochs, with a cap of 4 days, on 8 NVIDIA Tesla P40 GPUs, and Filipino and Oromo models for 40 epochs with a cap of 2 days on 4 NVIDIA Tesla P40 GPUs. We emphasize that the RoBERTa-MLM and RoBERTa-AG variants in an experiment are trained on the same number of epochs. We also pre-train both RoBERTa-12-MLM and RoBERTa-12-AG for longer and on TPUs to show that the trends hold even when using specialized hardware and more compute time (appendix ~\ref{appendix:longer}).
%Unlike state-of-the-art models, we note that our studies are performed on a smaller computational budget, both with respect to wall clock time, and hardware. Our English models are trained for 10 epochs, with a cap of 4 days, on 8 NVIDIA Tesla P40 GPUs, and Filipino and Oromo models for 40 epochs with a cap of 2 days on 4 NVIDIA Tesla P40 GPUs. However, we fully expect all trends to hold for larger models with more training epochs as well.

%%%\subsection{Auxiliary loss variants}
%%%
%%%There are two additional design choices for our approach when compared to RoBERTa-MLM — $\alpha$, and the attention patterns.
%%
%%%\%[  \mathcal{L} = \mathbb{E}_{t \sim \mathcal{D}}\left [\mathcal{L}_{MLM}(t) + \alpha \times \mathcal{L}_{attn}(t)\right ]\]
%%%
%%
% \note{Talk about dropout and warmup}
\section{Results} \label{sec:results}

%The proposed auxiliary loss improves the speed of convergence and makes the model more robust to hyperparameters. We show improvements on various domains.

% \begin{itemize}
%     \item A very small modification to existing models and code base
%     \item Low memory footprint, because the same attention pattern matrix can be reused in every layer
%     \item Minimal compute time addition ($15\%$ for smaller models, and negligible\note{get exact stats} for larger models)
%     \item Robust to hyperparameters like learning rate, batch size, attention dropout, and warmup steps unlike BERT and RoBERTa
%     \item Serves as a regularizer for an over-parameterized model
% \end{itemize}
\subsection{Language Modeling}

\begin{figure}[ht]
% \vskip 0.2in
\begin{center}
\centerline{\includegraphics[width=\columnwidth]{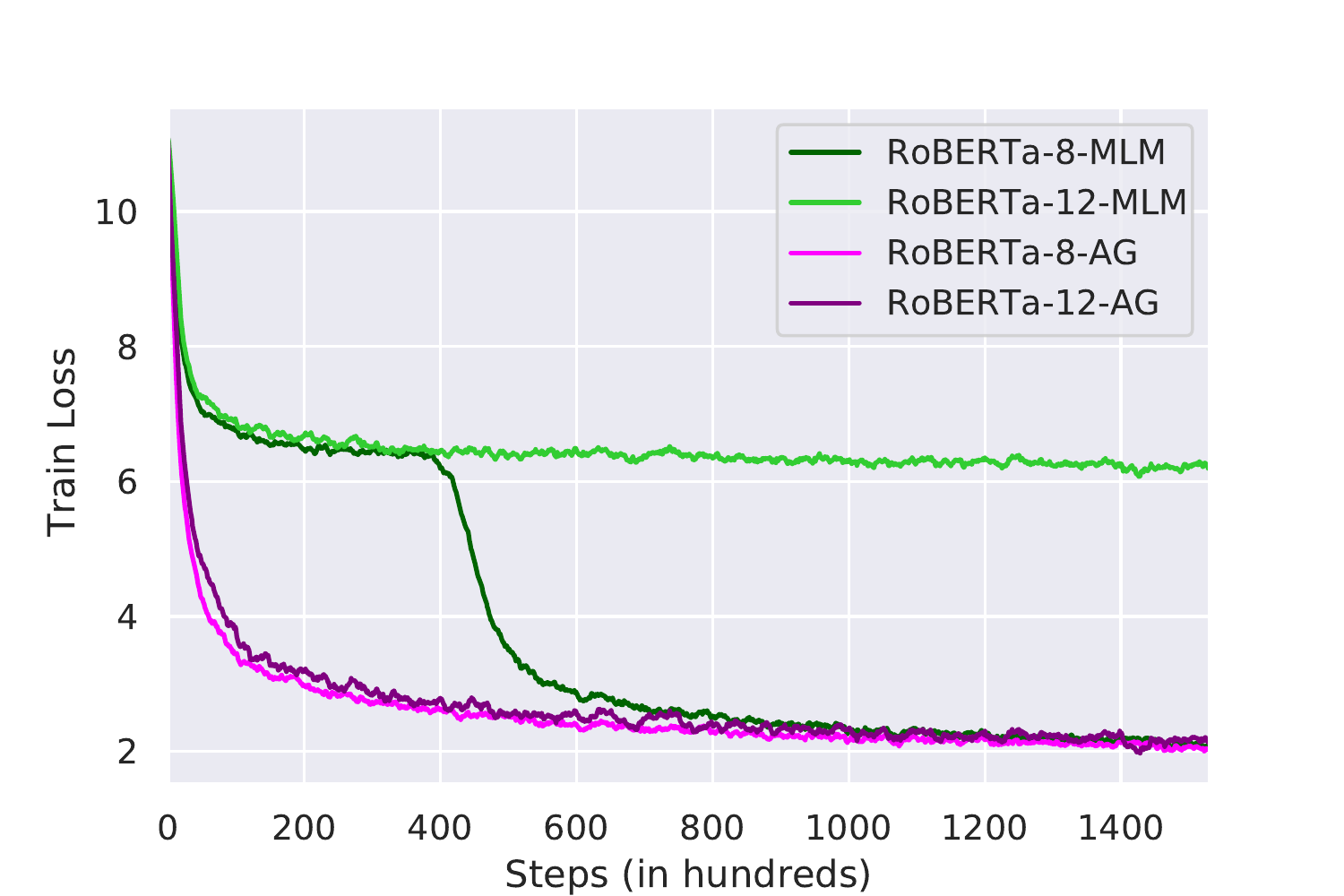}}
\caption{MLM Loss Curves for first 150k steps when training on English Wikipedia. Our AG models begin to converge instantly, while the MLM models have an extended plateau}
\label{fig:rob8rob12curves}
\end{center}
\vskip -0.2in
\end{figure}

\paragraph{Faster convergence}
Table~\ref{table:wiki_results} provides an overview of our results on language modeling.
As seen from the average loss, we observe that the AG loss greatly helps improve the speed of convergence on all model sizes and domains. Figure \ref{fig:rob8rob12curves} shows the train loss curves for two model sizes trained on English, where the losses for AG models almost instantaneously drop, whereas the MLM models have an extended period where the losses don't reduce. The gains are particularly notable for larger models like RoBERTa-12 and RoBERTa-16, where careful hyperparameter tuning is required for guaranteeing convergence if AG loss is not used. In contrast, using our auxiliary loss allows for fast convergence with standard out-of-the-box hyperparameters. For example, after just a day's training, the MLM loss for RoBERTa-16-AG has decreased from \textbf{11} to \textbf{2.5}, whereas RoBERTa-16-MLM's is still at \textbf{6.5}.

\paragraph{Final loss values}
Not only do the AG models converge faster, but their final train and validation losses are also lower than their MLM counterparts on 8 out of 9 settings (table \ref{table:wiki_results}). This is facilitated by AG models' fast initial convergence coupled with robustness to hyperparameters, allowing us to use larger learning rates and no warmup period. On 5 of the 9 settings, namely 12,16 layer models on Filipino and Oromo, and the 16 layer model on English, only our AG model can converge. We also provide a hypothesis about the usefulness of AG loss in appendix~\ref{appendix:aguseful}.

\begin{table*}[t]
\centering
% \resizebox{2\columnwidth}{!}{
\begin{tabular}{cccccccc}\toprule
\multirow{2}{*}{\textbf{Language}} & \multirow{2}{*}{\textbf{Loss}} & \multicolumn{2}{c }{\textbf{RoBERTa-8}} &\multicolumn{2}{c }{\textbf{RoBERTa-12}} & \multicolumn{2}{c }{\textbf{RoBERTa-16}} \\ \cmidrule(lr){3-4}  \cmidrule(lr){5-6}  \cmidrule(lr){7-8}  
%  & \multicolumn{2}{c }{\textbf{Train}}  & \multicolumn{2}{c }{\textbf{Validation}} & \multicolumn{2}{c }{\textbf{Train}}  & \multicolumn{2}{c }{\textbf{Validation}}& \multicolumn{2}{c }{\textbf{Train}}  & \multicolumn{2}{c }{\textbf{Validation}}\\ \cmidrule(lr){2-3} \cmidrule(lr){4-5}  \cmidrule(lr){6-7} \cmidrule(lr){8-9} \cmidrule(lr){10-11} \cmidrule(lr){12-13}
 & & MLM & AG & MLM & AG  &  MLM & AG \\ \midrule
%  &  MLM & AG & MLM & AG  &  MLM & AG & MLM & AG  &  MLM & AG & MLM & AG \\ \midrule
\multirow{3}{*}{\textbf{English}} & Train & 1.75 & \textbf{1.74} &1.86 &\textbf{1.73} &6.40&\textbf{1.81} \\
& Validation & \textbf{2.41} & 2.43 & 2.47 & \textbf{2.29} & 7.28 & \textbf{2.46} \\
& Average & 2.48 & \textbf{2.09} & 2.56 & \textbf{2.07} & 6.67 & \textbf{2.24} \\ \midrule
\multirow{3}{*}{\textbf{Filipino}} & Train & 3.10 & \textbf{0.74}  & 5.04 & \textbf{0.66} & 5.18 & \textbf{0.63} \\
& Validation & 3.22 & \textbf{0.99} & 5.06 & \textbf{0.97} & 5.05 & \textbf{0.82} \\
& Average & 4.95 & \textbf{1.20} & 5.24 & \textbf{1.16} & 5.28 & \textbf{1.11} \\ \midrule
\multirow{3}{*}{\textbf{Oromo}} & Train & 3.44 & \textbf{3.31} & 6.24 & \textbf{3.23} & 6.51 & \textbf{3.34} \\
& Validation & 4.06 & \textbf{3.93} & 6.74 & \textbf{3.88} & 6.93 & \textbf{3.92} \\
& Average & 5.15 & \textbf{4.52} & 6.75 & \textbf{4.52} & 6.95 & \textbf{4.62} \\
\bottomrule
\end{tabular}
% }
\caption{Train, validation, and average train MLM loss on all three languages. Even after 4 days of training (section~\ref{section:implementation}), our AG models outperform MLM models on all but one settings. Comparisons are columnwise.}
\label{table:wiki_results}
\end{table*}

\subsection{Downstream performance}

We evaluate all the models' downstream performance to verify if better language modeling corresponds to better language understanding.

\paragraph{English}

Our AG models outperform their MLM counterparts on 11 out of the 12 settings (Table \ref{table:downstream}), with 7 comparisons being statistically significant ($p<.05$, paired t-test). We emphasize that the scores are not directly comparable to RoBERTa$_{\textrm{BASE}}$, which is trained on $10$ times more data, up to $8$ times more epochs, and on several GPUs. Having experimentally shown the usefulness of AG loss in optimizing the MLM objective, we believe that training our models on more data and compute is bound to match or outperform the MLM counterparts. 

\paragraph{Filipino} As shown in table \ref{table:downstream}, our AG models outperform the MLM variants on all model sizes. Additionally, our best performing model beats the current SOTA~\cite{cruz2019evaluating} by almost $1$ point, even though the latter was trained on a TPU and for longer wall-clock time.

\paragraph{Oromo} Our AG models continue to have an edge even in sparse data domains. Though Oromo has only 0.2\% of the pre-training data when compared to English, which makes it prone to overfitting, it is interesting to note that larger models continue to outperform smaller ones on downstream tasks. 
% We speculate that this can partly be attributed to the regularizing effect of guiding attention, which constraints the parameters\ameet{Speculative}. 
Our models are competitive with E-MBERT~\cite{wang2020extending}, which is a BERT model leveraging resources from over 104 other languages. We hope that our competitive results on both Filipino and Oromo when using as little as 4 GPUs encourages more NLP research in low-resource languages.

\begin{table*}[t]
\centering
\resizebox{\linewidth}{!}{
\begin{tabular}{ccllllllcc}\toprule
\multirow{2}{*}{\textbf{Language}} & \multirow{2}{*}{\textbf{Task}} & \multicolumn{2}{c }{\textbf{RoBERTa-8}} &\multicolumn{2}{c }{\textbf{RoBERTa-12}} & \multicolumn{2}{c }{\textbf{RoBERTa-16}} & \multicolumn{2}{c }{\textbf{SOTA}} \\ \cmidrule(lr){3-4}  \cmidrule(lr){5-6}  \cmidrule(lr){7-8} \cmidrule(lr){9-10}  
%  & \multicolumn{2}{c }{\textbf{Train}}  & \multicolumn{2}{c }{\textbf{Validation}} & \multicolumn{2}{c }{\textbf{Train}}  & \multicolumn{2}{c }{\textbf{Validation}}& \multicolumn{2}{c }{\textbf{Train}}  & \multicolumn{2}{c }{\textbf{Validation}}\\ \cmidrule(lr){2-3} \cmidrule(lr){4-5}  \cmidrule(lr){6-7} \cmidrule(lr){8-9} \cmidrule(lr){10-11} \cmidrule(lr){12-13}
 & & MLM & AG & MLM & AG  &  MLM & AG & Model & Score \\ \midrule
%  &  MLM & AG & MLM & AG  &  MLM & AG & MLM & AG  &  MLM & AG & MLM & AG \\ \midrule
\multirow{4}{*}{\textbf{English}} & MNLI-m & 78.8 & \textbf{79.1} & 78.9 & \textbf{79.0} & 69.7 & \textbf{79.6}$^+$ & \multirow{4}{*}{RoBERTa$_{\textrm{BASE}}$~\citep{liu2019roberta}} & 87.6$^\star$ \\
& MNLI-mm & 77.6 & \textbf{77.7} & 77.6 & \textbf{78.9}$^+$ & 68.8 & \textbf{78.7}$^+$ &  &87.6$^\star$ \\
& QNLI & 84.3 & \textbf{84.6} & 86.1 & \textbf{86.8} & 72.0 & \textbf{84.4}$^+$ &  & 92.8$^\star$ \\
& QQP & \textbf{69.1} & 68.3$^+$ & 68.4 & \textbf{68.9}$^+$ & 58.2 & \textbf{68.5}$^+$ &  & 74.3$^\star$ \\ \midrule
\multirow{1}{*}{\textbf{Filipono}} & Sentiment & 74.1 & \textbf{75.6}$^{\star +}$ & 74.1 & \textbf{74.6} & 74.1 & \textbf{75.5}$^+$ & BERT~\citep{cruz2020establishing} & 74.8 \\ \midrule
\multirow{1}{*}{\textbf{Oromo}} & NER & 64.6 & \textbf{66.7}$^+$ & 51.5 & \textbf{67.9}$^+$ & 53.5 & \textbf{67.2}$^+$ & E-MBERT~\citep{wang2020extending} & 72.8$^\star$  \\
\bottomrule
\end{tabular}
}
\caption{Evaluation on downstream tasks. Our AG models outperform their MLM counterparts on all but one settings (entries marked with `$+$` are significant with $p<.05$, paired t-test). Comparisons are column-wise. SOTA=state-of-the-art published numbers (marked $^{\star}$) with similar model types on each task. The SOTA models are trained on more compute and data and are not directly comparable to our models.}
\label{table:downstream}
\end{table*}

\subsection{Ablation study with attention patterns} \label{appendix:pattern_ablation}
As mentioned in section~\ref{section:method}, we introduce five different attention patterns for guiding our models using the AG loss. To select the best performing patterns, we use the leave-one-out strategy, in which we omit patterns and record the increases in loss (after 100,000 steps) when compared to a model with all patterns included. The patterns which cause a large increase in loss when omitted are naturally more important. The increases in loss are recorded in Table~\ref{appendix:table:robustness}, which shows that \texttt{[Next,Prev]} patterns are most important, followed by \texttt{[First]} and \texttt{[Period]}, while \texttt{[Delim]} isn't very useful. Furthermore, unlike \texttt{[Period]}, the \texttt{[First]} pattern's guidance matrix $\mathbf{P}$ (section~\ref{section:method}) is fixed, making it more computationally efficient to use. Hence, we guide one head each with \texttt{[Next,Prev]} patterns, and $(\lambda h-2)$ heads with the \texttt{[First]} pattern.

\begin{table}[t]
\centering
% \resizebox{\columnwidth}{!}{
\begin{tabular}{c c}
\toprule
\textbf{Pattern(s) omitted} & \textbf{Change in loss} \\ \midrule 

\texttt{[Next,Prev]} & $2.99\to 7.74$ \\
\texttt{[First]} & $2.99\to 3.02$ \\
\texttt{[Period]} & $2.99\to 3.03$ \\
\texttt{[Delim]} & $2.99\to 2.99$ \\
\bottomrule
\end{tabular}
% }
\caption{Ablation study for choosing the best performing attention patterns to use for guidance (AG loss). The entry $x\to y$ means that the loss after omitting the respective pattern increased from $x$ to $y$. We see that \texttt{[Next],[Prev]} are most important followed by \texttt{[First]} and \texttt{[Period]}.}
\label{appendix:table:robustness}
\end{table}
% We use the leave-one-out strategy where patterns which cause larger increases in loss when omitted are more important.

\subsection{Attention Guidance for ELECTRA}

ELECTRA~\cite{clark2020electra} is an efficient model which uses \textit{replaced token detection} as the pre-training task. It comprises training a discriminator and a generator, in which the generator randomly changes $k\%$ of tokens in an input sequence to plausible alternatives, and the discriminator has to identify if a token was modified or not. The generator learns using the MLM objective, and the discriminator, which is used for downstream tasks, uses the logistic loss. We use an ELECTRA variant in which the generator is a unigram LM, and compare the performance when AG loss is added. The results after training ELECTRA-12 and ELECTRA-12-AG for $2$ epochs on BooksCorpus~\cite{zhu2015aligning} are presented in Table \ref{table:electra}. Like with RoBERTa, we report only the discriminator's logistic loss even though our model is trained on an auxiliary loss. The AG model shows gains in convergence without any ELECTRA specific hyperparameter tuning.%, and corroborates the hypothesis that AG loss can be useful for multiple self-supervised training objectives.

\begin{table}
\centering
\begin{tabular}{c c c}
\toprule
\textbf{Loss} & \textbf{ELECTRA} & \textbf{ELECTRA-AG} \\ \midrule 

\textbf{Final} & 0.27  & \textbf{0.10} \\
\textbf{Average} & 0.32  & \textbf{0.15} \\ \bottomrule
\end{tabular}
\caption{Training loss and average training loss over two epochs for ELECTRA discriminator. Adding AG loss reduces both the final and average training loss}
\label{table:electra}
\end{table}

\subsection{Attention Guidance for Machine Translation}\label{results:mt}

% \lipsum[1]

\paragraph{Models} We also experiment with adding our AG loss to Machine Translation (MT) models that use Transformers for both the encoder and decoder. 
We compare with the BASE Transformer~\cite{vaswani2017attention} and a recently proposed hard-coded Gaussian model~\cite{you2020hard}, which fixes all the attention heads in the encoder and decoder to pre-determined Gaussian distributions centered around nearby tokens.
While the latter's attention patterns are similar to our \textit{local} attention patterns, they are hard-coded and not an auxiliary loss.
Following~\cite{you2020hard}, the cross-attention in our MT model is not guided.
Using a held-out set, we search for the best combination of AG patterns (Figure~\ref{fig:attentionpatterns}) for both the encoder and decoder. We find this to be one head each guided with the \texttt{[Next,Prev]} pattern in the encoder, and no heads being guided in the decoder. \textbf{Global} patterns (like attending to \texttt{[First]}) seem to be detrimental to performance in MT.

\paragraph{Results} We perform experiments on IWSLT16 En-De~\cite{cettolo2016iwslt} and WMT14 En-De datasets, and report train negative log-likelihood (NLL), validation NLL, average train NLL (to compare convergence speed), and the BLEU score on the test set. All models are trained for $100,000$ steps. Similar to  LM pre-training, we observe that our model has the lowest train, validation, and average NLL for both the datasets, showing that guiding attention heads helps even with MT. Furthermore, the AG model's BLEU scores are comparable to the scores of BASE and hard-coded Gaussian. We note that our AG patterns are tailored for language-modeling, and MT models could benefit from a more extensive search over possible patterns.

\begin{table}
\centering
\resizebox{\columnwidth}{!}{
\begin{tabular}{clccc}\toprule
\textbf{Dataset} & \textbf{Loss/Metric} & \textbf{BASE} &\textbf{Hard-coded} & \textbf{AG} \\ \midrule %\cmidrule(lr){3-4}  \cmidrule(lr){5-6}  \cmidrule(lr){7-8}
%  & & MLM & AG & MLM & AG  &  MLM & AG \\ \midrule
%  &  MLM & AG & MLM & AG  &  MLM & AG & MLM & AG  &  MLM & AG & MLM & AG \\ \midrule
%----------- IWSLT ---------------------------
\multirow{4}{*}{\textbf{IWSLT}} & Train NLL & \textbf{1.18} & 1.30 & \textbf{1.18} \\
 & Average NLL & 1.88 & 1.92 & \textbf{1.84} \\
 & Validation NLL & 2.25 & 2.26 & \textbf{2.22} \\
 & BLEU & \cmmnt{29.90,}\textbf{24.52}  & \cmmnt{29.93,}24.42 & \cmmnt{27.90,}24.42 \\ \midrule
 %----------- WMT ----------------------------
\multirow{4}{*}{\textbf{WMT}} & Train NLL & 1.77 & 1.94 & \textbf{1.75} \\
 & Average NLL & 2.07 & 2.23 & \textbf{2.05}\\
 & Validation NLL & 1.61 & 1.73 & \textbf{1.60} \\
 & BLEU & \textbf{26.24} & 25.50 & 23.34 \\ \bottomrule
\end{tabular}
}
\caption{Comparing the train, average train, and validation negative log-likelihood (NLL) loss, and also the BLEU scores for BASE (standard Transformer), Hard-Coded~\cite{you2020hard}, and AG (our model). AG model has the lowest NLL losses and its BLEU scores are comparable to the other models.}
\label{table:stupidnmt}
\end{table}

\subsection{Probing analysis}
% \karthik{should we swap this section with Electra? Since we use the Roberta models here and not Electra?}
% \note{Should we add structure of attention in different language points here?}

\begin{table*}
\centering
\resizebox{\linewidth}{!}{
\begin{tabular}{l c c c c c c}
\toprule
\multirow{2}{*}{\textbf{Model}} & \textbf{MLM loss} & \multicolumn{4}{c}{\textbf{CoNLL-2012}} & \multicolumn{1}{c}{\textbf{Synthetic}} \\ \cmidrule(lr){3-6} \cmidrule(lr){7-7}
 & \textbf{(valid)} & \textbf{ALL} & \textbf{NOMINAL} & \textbf{PRONOMINAL} & \textbf{PROPER} & \textbf{ALL}\\ \midrule 
% & \textbf{Max acc} & \textbf{Mean acc} \\ \midrule 

Rule-Based & - & 0.66 & 0.48 & 0.72 & 0.73 & -\\
Randomly Initialized & 11.0 & 0.50 & 0.41 & 0.47 & 0.60 & 26.6\\ \cdashlinelr{1-7}
BERT$_{\textrm{BASE}}$~\citep{devlin2018bert} & - & 0.70 & 0.64 & 0.68 & 0.76 & 0.97\\

RoBERTa$_{\textrm{BASE}}$~\citep{liu2019roberta} & - & 0.74 & 0.71 & 0.74 & 0.76 & 0.99\\ \cdashlinelr{1-7}

% RoBERTa$_{\textrm{BASE}}$~\cite{liu2019roberta} & - & \textbf{99.8} & 6.7 \\
% RoBERTa-Attn ($\lambda=1/2$) & 0.674 &  0.047 & 0.704 & 0.047 \\
RoBERTa-MLM & 2.47 & 0.68  & 0.58 & 0.69 & 0.73 & 0.87\\
% RoBERTa-MLM (Ours) & 2.47 & 86.8  & 12.8   \\
RoBERTa-AG ($\lambda=1/2$) & 2.29 & 0.28  & 0.21  &  0.32 & 0.29 & 0.84\\
% RoBERTa-AG ($\lambda=1/2$) & 2.29 & 83.6  & 4.7  \\
RoBERTa-AG ($\lambda=1$) & 2.31 & \textbf{0.21}  & \textbf{0.13} & \textbf{0.21}  & \textbf{0.28} & \textbf{0.00}\\ \bottomrule
% RoBERTa-AG ($\lambda=1$) & 2.31 & \textbf{0.0} & 0.0 \\
% Randomly Initialized & - & 26.6 & 15.4 \\ \bottomrule
\end{tabular}
}
\caption{Probing analysis to measure coreference resolution accuracies of the best performing attention head from each model on CoNLL-2012~\cite{clark2019does} and synthetic~\cite{lin2019open} datasets. Interestingly, our AG models (last two rows) can be better at language modeling (lower MLM loss) without having a single head that is good at coreference. (\textbf{ALL}=overall mean scores, \textbf{Bold=lowest})} 
\label{table:coreference}
\end{table*}

Motivated by recent studies \cite{clark2019does,lin2019open,Manning201907367} which posit that individual attention heads can encode linguistic information, we analyze attention patterns in the self-attention heads of our models. Specifically, we search for heads that can individually perform coreference resolution.

\paragraph{Method} We use the probe described in \cite{clark2019does}, which evaluates attention heads on antecedent selection accuracy. A sentence (e.g. ``\textit{\textbf{The CEO} led \textbf{her} company to success}") is input to the model, and each head is scored on its ability to identify antecedents, e.g. a score of 1 if the token `\textit{her}' attends most to a token in `\textit{The CEO}'. We aggregate the scores over all the coreferent mention-antecedent pairs in the dataset and report the accuracy of each model's best performing head. We also include the scores of a \textit{randomly initialized} RoBERTa model for comparison. We leave further details to~\citet{clark2019does}.

\paragraph{Datasets} Following~\citet{clark2019does}, we evaluate our models on the CoNLL-2012 dataset~\cite{pradhan2012conll}. We also evaluate on a synthetic dataset of $10000$ samples from~\citet{lin2019open} and follow their method of adding a distractor sentence (e.g. adding ``\textit{The \textbf{people} were happy}" after ``\textit{\textbf{The CEO} led \textbf{her} company to success}") which serves to introduce spurious entities. We ensure that the antecedent is not the word directly before the coreferent mention so that a trivial baseline which always chooses the previous word gets a score of $0$.

% \begin{table*}
% \centering
% \resizebox{\linewidth}{!}{
% \begin{tabular}{l c c c}
% \toprule
% \textbf{Model} & \textbf{Valid. Loss} & \textbf{Max acc} & \textbf{Mean acc} \\ \midrule 

% RoBERTa$_{\textrm{BASE}}$~\cite{liu2019roberta} & - & \textbf{99.8} & 6.7 \\
% % RoBERTa-Attn ($\lambda=1/2$) & 0.674 &  0.047 & 0.704 & 0.047 \\
% RoBERTa-MLM (Ours) & 2.47 & 86.8  & 12.8   \\
% RoBERTa-AG ($\lambda=1/2$) & 2.29 & 83.6  & 4.7  \\
% RoBERTa-AG ($\lambda=1$) & 2.31 & \textbf{0.0} & 0.0 \\
% Randomly Initialized & - & 26.6 & 15.4 \\ \bottomrule
% \end{tabular}
% }
% \caption{Coreference Resolution accuracies. \textbf{Max acc}: accuracy of best performing head, \textbf{Mean acc}: average accuracy over all heads. RoBERTa-AG (fourth row) has no head which performs well while still being better than the baseline (second row) at language modeling}
% \label{table:coreference}
% \end{table*}

\paragraph{Discussion}  We discuss results reported in Table \ref{table:coreference}. We observe the same trends on both the CoNLL-2012 dataset and the synthetic dataset and discuss the former in detail. In line with~\citet{clark2019does}'s observation, BERT and RoBERTa have heads which achieve the highest accuracies. Even though RoBERTa-MLM (section ~\ref{sec:model_definitions}) is trained on significantly lesser compute and data, its performance is comparable to BERT and better than the Rule-based baseline. But interestingly, both RoBERTa-AG ($\lambda=1/2$) and RoBERTa-AG ($\lambda=1$), which have half and all their heads guided respectively, perform significantly worse than both the baseline and a randomly initialized (untrained) model. Surprisingly, this is true even though the validation loss for both RoBERTa-AG ($\lambda=1/2$) and RoBERTa-AG ($\lambda=1$) is lower (better) than RoBERTa-MLM's. The performance degradation in RoBERTa-AG models is because half/all the heads pay most of their attention to a predefined pattern, thus rendering them unable to pay attention to the antecedent. This provides evidence that language modeling performance is not necessarily correlated with the performance of individual heads on linguistic tasks, and that attention patterns of the heads are not necessarily directly interpretable. This observation is in line with a recent study~\citep{brunner2020identifiability} that questions the interpretability of attention distributions.

The trends on the synthetic dataset (Table~\ref{table:coreference}) are similar where BERT and RoBERTa have a head that achieves close to perfect accuracy, and RoBERTa-MLM has a head whose accuracy is significantly better than that of a randomly initialized model. However, RoBERTa-AG ($\lambda=1$) performs poorly (an accuracy of $0$) even though its validation MLM loss is lower (better) than RoBERTa-MLM's.

% In line with~\cite{clark2019does}'s observation, BERT and RoBERTa have a head that achieves close to perfect accuracy. Similarly, both RoBERTa-MLM and RoBERTa-AG ($\lambda=1/2$), which have half the heads guided, have a head whose accuracy is significantly better than that of a randomly initialized (RI) model. However, RoBERTa-AG ($\lambda=1$), a variant of our AG model which guides all the heads and performs only negligibly worse than RoBERTa-AG ($\lambda=1/2$) on language modeling, performs poorly on coreference resolution. In fact, even though its validation loss is lower than RoBERTa-MLM's, it does worse than a randomly initialized model at co-reference resolution.  This is because each head pays most of it's attention to a predefined pattern, thus rendering it unable to pay attention to the antecedent. This gives some evidence that individual heads need not be performing linguistically motivated tasks even if the model as a whole performs well at language modeling and that the attention patterns of the heads are not necessarily directly interpretable. This observation is in line with a recent study~\cite{brunner2020identifiability} which questions the interpretability of attention distributions.
\section{Conclusion}\label{sec:conclusions}

In this study, we introduce the simple yet effective Attention Guidance (AG) loss, which speeds up convergence and improves performance on various domains and model sizes. Adding this loss also makes Transformers robust to hyperparameters like learning rate, warmup steps, and dropout. Our experiments also show its usefulness in multiple pre-training objectives. The gains are particularly strong on larger models, enabling their usage in low-compute scenarios and low-resource domains. Our analysis of the relation of AG loss and MLM loss shows the usefulness of our method, and we hope that this paper can serve as a starting point for future works aiming to exploit and question self-attention in Transformers.

\section*{Acknowledgement}
This research was partially funded by the Center for Statistics and Machine Learning at Princeton University through support from Microsoft. This work was also supported with Cloud TPUs from Google's TensorFlow Research Cloud (TFRC). We thank Austin Wang, Jens Tuyls, Zexuan Zhong, Michael Hu, and Danqi Chen for providing valuable comments and feedback.

\newpage

\bibliographystyle{acl_natbib}
\bibliography{emnlp2020}

% Submit appendix.pdf as a separate pdf file
\appendix
\section{Appendices} \label{sec:appendix}

\subsection{Glossary}

\begin{itemize}
    \item AG: Attention Guidance
    \item AG Loss: Attention Guidance Loss
    \item AG Model: Attention Guided Model
    \item RoBERTa-$X$-AG: An $X$ layer RoBERTa model trained with MLM and AG loss
    \item MLM: Masked Language Modeling
    \item RoBERTa-$X$-MLM: An $X$ layer RoBERTa model trained with only MLM loss
    \item SOTA: State-of-the-art
    \item Head: Self-Attention Heads
\end{itemize}

\subsection{Mathematical Specification of Patterns}\label{appendix:mathspec}

Please refer to section \ref{sec:approach} for definitions.\\

Let $CNT($`\texttt{token}'$)$ be the total number of occurences of \texttt{token} in an input $I$ of length $n$. $I[j]$ represents the $j^{th}$ token in $I$. Let \texttt{DELIM} represent the set of all delimiters added by the tokenizer.

\begin{equation*}
    \begin{split}
        \mathbf{P}_{\textrm{\texttt{[Next]}}}[p,q] =& \begin{cases} 1 &\mbox{if } q = p+1 \\
    \frac{1}{n} &\mbox{if } p = n \\
    0 & \textrm{otherwise} \end{cases}\\
        \mathbf{P}_{\textrm{\texttt{[Prev]}}}[p,q] =& \begin{cases} 1 &\mbox{if } q = p-1 \\
    \frac{1}{n} &\mbox{if } p = 1 \\
    0 & \textrm{otherwise} \end{cases}\\    
        \mathbf{P}_{\textrm{\texttt{[First]}}}[p,q] =& \begin{cases} 1 &\mbox{if } q = 1 \\ 
0 & \textrm{otherwise} \end{cases}\\ 
        \mathbf{P}_{\textrm{\texttt{[Period]}}}[p,q] =& \begin{cases} \frac{1}{CNT(\textrm{`\texttt{.}'})} &\mbox{if } I[q] = `\textrm{\texttt{.}'}\\ 
0 & \textrm{otherwise} \end{cases}\\ 
        \mathbf{P}_{\textrm{\texttt{[Delim]}}}[p,q] =& \begin{cases} \frac{1}{CNT(\textrm{`\texttt{DELIM}'})} &\mbox{if } I[q] \in \textrm{\texttt{DELIM}}\\ 
0 & \textrm{otherwise} \end{cases}\\ 
    \end{split}
\end{equation*}

\begin{figure*}[ht]
% \vskip 0.2in
\begin{center}
\centerline{\includegraphics[width=\linewidth]{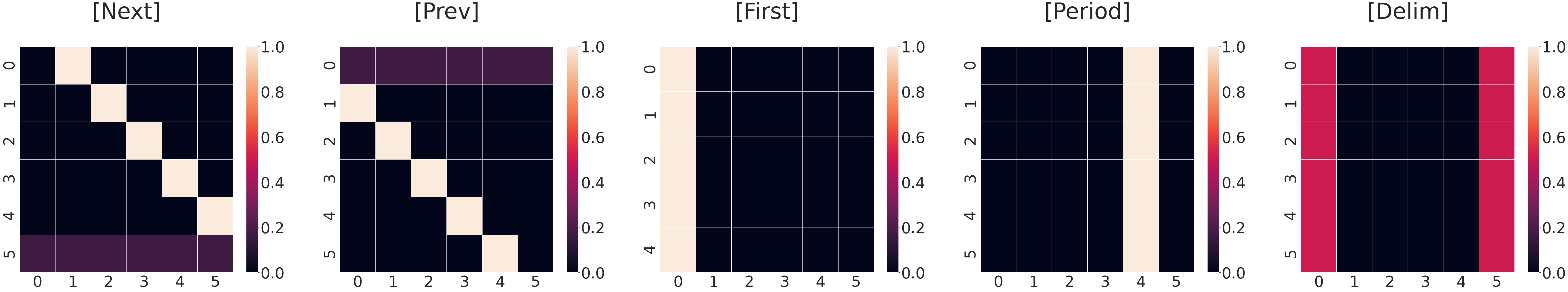}}
\caption{Example attention patterns for the sentence ``\textit{$<$s$>$ Welcome to EMNLP . $</$s$>$}". Note that the first three patterns don't depend on the sentence, and can be considered fixed patterns, and the last two depend on the position of the period and delimiters respectively.}
\label{appendix:fig:attentionpatterns}
\end{center}
\vskip -0.2in
\end{figure*}

We add the patterns in figure~\ref{appendix:fig:attentionpatterns} for reference.

\subsection{Why is AG Loss Useful?} \label{appendix:aguseful}
\begin{figure}[H]
% \vskip 0.2in
\begin{center}
\centerline{\includegraphics[width=\columnwidth]{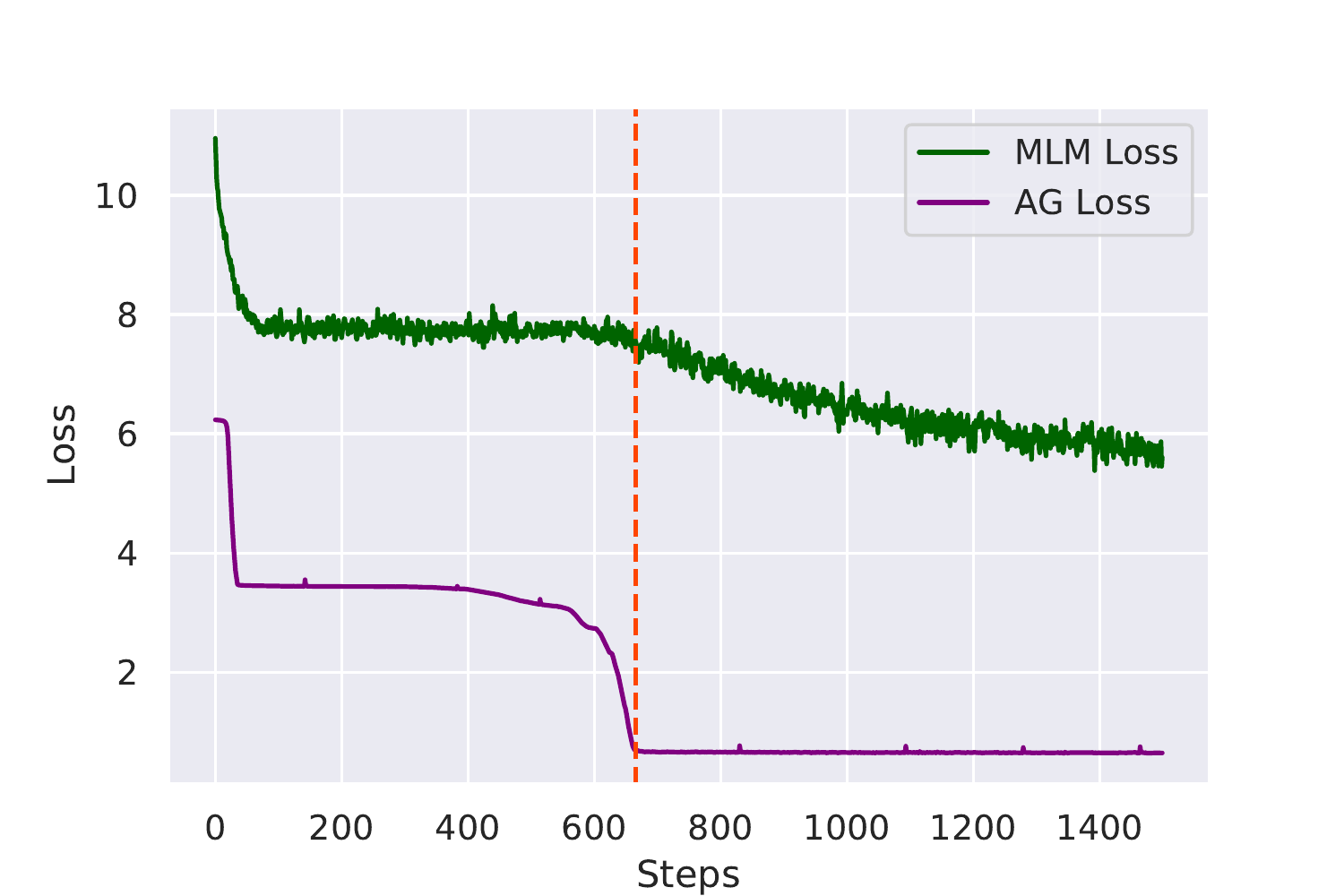}}
\caption{MLM loss ($\mathcal{L}_{MLM}$) and AG loss ($\mathcal{L}_{AG}$) for RoBERTa-12-AG. The MLM loss starts dropping as soon as AG loss converges.}
\label{fig:mlmattncorr}
\end{center}
% \vskip -0.4in
\end{figure}

The AG loss converges within 0.2\% of pre-training time. This fast convergence is because it is simple to attend to our patterns, which only require propagation of the positional embedding (for \texttt{[Next]}, \texttt{[Prev]}), or the non-contextual word embeddings in the input layer (for \texttt{[Delim]}, \texttt{[Period]}). In theory, this is particularly easy for Transformers because of the presence of residual connections~\cite{he2016deep}. We observe from Figure~\ref{fig:mlmattncorr} that as soon as AG loss converges, the MLM loss starts decreasing, and we hypothesize that the quick convergence of AG loss because of the reasons explained above is responsible for our method's advantages.

% \vskip -0.1in

\subsection{Running English RoBERTa models on TPUs}\label{appendix:longer}

To show that the trends mentioned in Table~\ref{table:downstream} hold even when specialized hardware is used, and models are run for longer, we run the RoBERTa-12-AG and RoBERTa-12-MLM models for 8 epochs (as opposed to 7) on TPUs. The results in Table~\ref{table:longerrunning} show that AG models to continue to have the advantage over MLM models.

\begin{table*}[t]
\centering
\resizebox{\linewidth}{!}{
\begin{tabular}{ccccccc}\toprule
 \multirow{2}{*}{\textbf{Task}} & \multicolumn{2}{c }{\textbf{RoBERTa-12-GPU}} & \multicolumn{2}{c }{\textbf{RoBERTa-12-TPU}} & \multicolumn{2}{c }{\textbf{SOTA}} \\ \cmidrule(lr){2-3}  \cmidrule(lr){4-5}  \cmidrule(lr){6-7}   
%  & \multicolumn{2}{c }{\textbf{Train}}  & \multicolumn{2}{c }{\textbf{Validation}} & \multicolumn{2}{c }{\textbf{Train}}  & \multicolumn{2}{c }{\textbf{Validation}}& \multicolumn{2}{c }{\textbf{Train}}  & \multicolumn{2}{c }{\textbf{Validation}}\\ \cmidrule(lr){2-3} \cmidrule(lr){4-5}  \cmidrule(lr){6-7} \cmidrule(lr){8-9} \cmidrule(lr){10-11} \cmidrule(lr){12-13}
 & MLM & AG  &  MLM & AG & Model & Score \\ \midrule
%  &  MLM & AG & MLM & AG  &  MLM & AG & MLM & AG  &  MLM & AG & MLM & AG \\ \midrule
MNLI-m &  78.9 & \textbf{79.0} & 80.3 & \textbf{81.2} & \multirow{4}{*}{BERT$_{\textrm{BASE}}$~\cite{devlin2018bert}} & 84.6$^\star$ \\
MNLI-mm &  77.6 & \textbf{78.9} & 80.8 & \textbf{81.4} &  &83.4$^\star$ \\
QNLI &  86.1 & \textbf{86.8} & 88.7 & \textbf{89.0} &  & 90.5$^\star$ \\
QQP & 68.4 & \textbf{68.9} & 69.3 & \textbf{69.7} &  & 71.2$^\star$ \\
% MRPC & - & - & - & - & 87.9/82.8 & \textbf{90.8/87.2} &  & 88.9/84.8$^\star$ \\
% SST-2 & - & - & - & - & 91.0 & \textbf{91.1} &  & 93.5$^\star$ \\
% STS-B & - & - & - & - & 86.5/86.3 & \textbf{87.2/87.0} &  & 88.9/84.8$^\star$ \\
% WNLI & - & - & - & - & \textbf{23.9} & 12.6 &  & 65.1$^\star$ \\
% CoLA & - & - & - & - & \textbf{48.9} & 48.4 &  & 52.1$^\star$ \\
% RTE & - & - & - & - & 63.5 & \textbf{63.8} &  & 66.4$^\star$ \\
\bottomrule
\end{tabular}
}
\caption{RoBERTa-12-AG continues to outperform RoBERTa-MLM-AG even when trained on TPUs. This shows that using AG loss provides performance improvements even when using specialized hardware. We also report BERT-Base's scores for reference. Note BERT's scores are not directly comparable because it is trained for 40 epochs and our models are trained for 8.}
\label{table:longerrunning}
\end{table*}

\subsection{Train Loss Curves for ELECTRA}

\begin{figure}[H]
% \vskip 0.2in
\begin{center}
\centerline{\includegraphics[width=\columnwidth]{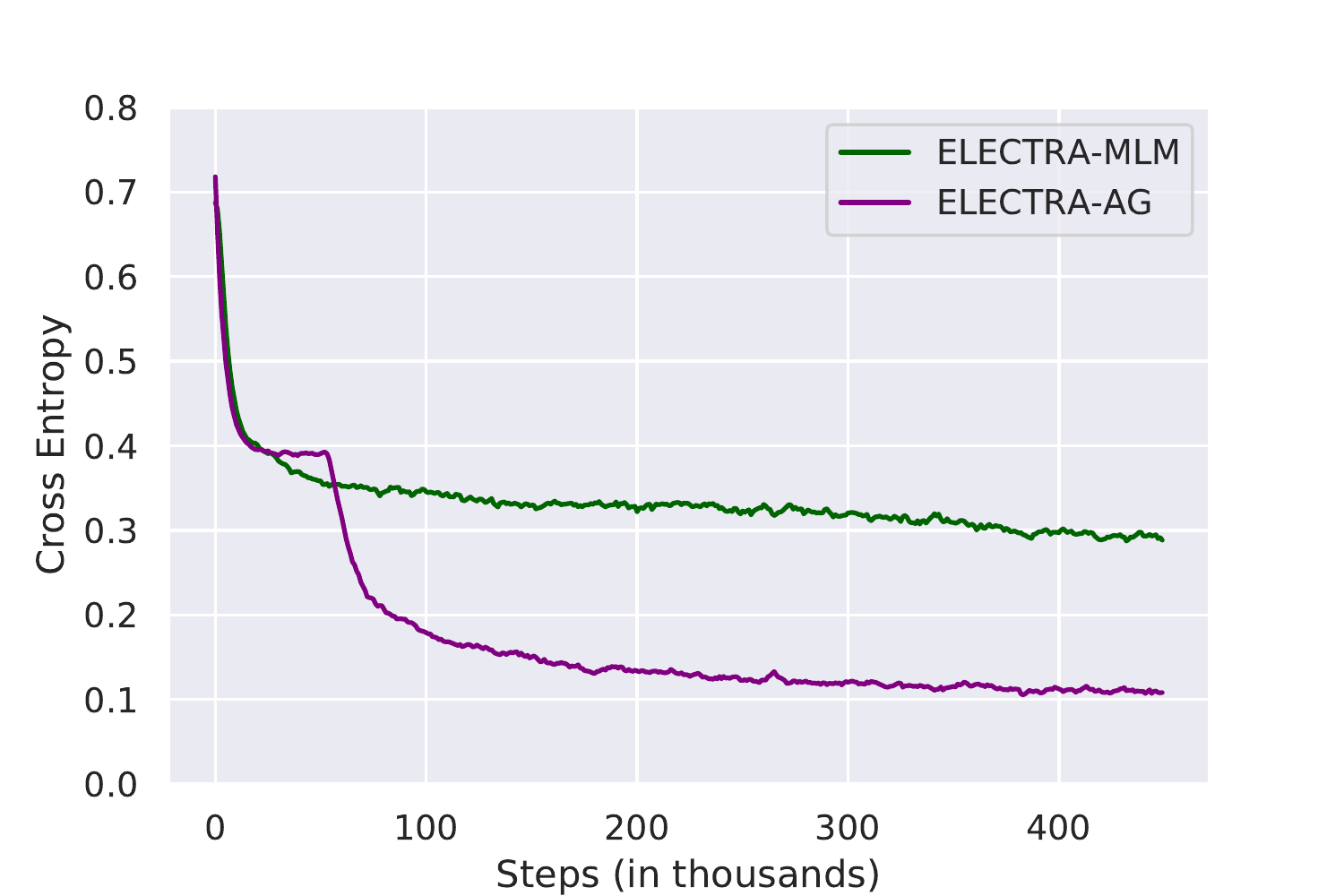}}
\caption{Loss Curves for ELECTRA. The MLM model has an extended plateau, whereas our AG model starts converging almost instantly.}
\label{appendix:fig:electra}
\end{center}
\end{figure}

Our method shows faster convergence with ELECTRA, as shown in Figure~\ref{appendix:fig:electra}.

\subsection{Train Negative Log-Likelihood Curves for Machine Translation}

\begin{figure*}
\begin{subfigure}{.45\textwidth}
  \centering
  \includegraphics[width=\linewidth]{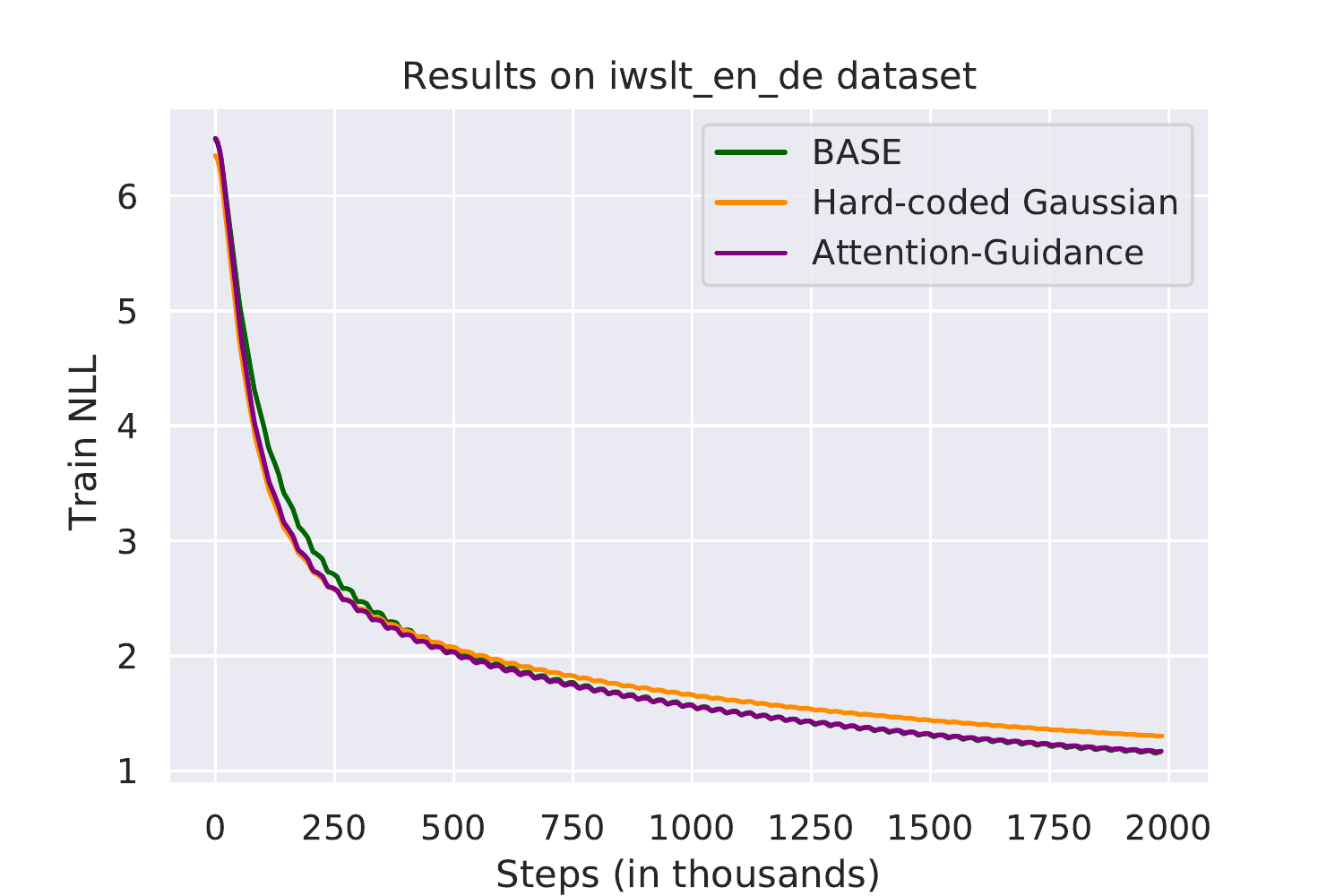}
%   \caption{Sing-D}\vspace{1em}
\end{subfigure}\hfill
\begin{subfigure}{.45\textwidth}
  \centering
  \includegraphics[width=\linewidth]{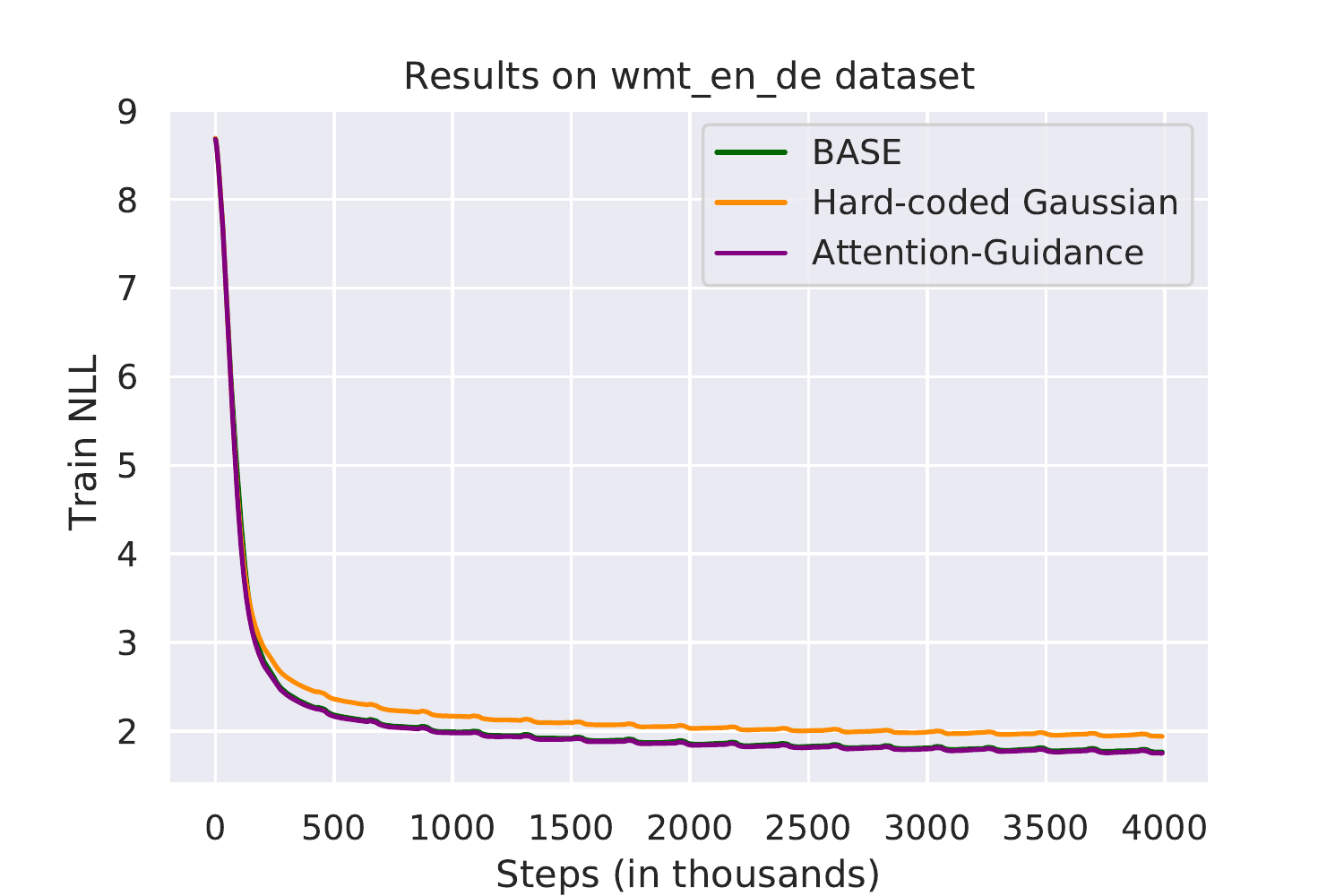}
%   \caption{Dual-D}
\end{subfigure}
\caption{Train loss curves on IWSLT-en-de dataset (left) and WMT-en-de dataset (right). Our AG model converges to the same loss as the BASE model, but the Hard-coded Gaussian model~\cite{you2020hard} converges to a higher loss.}
\label{appendix:fig:mtcurves}
\end{figure*}

We report the train loss curves for experiments on machine-translation (section ~\ref{results:mt}) in Figure~\ref{appendix:fig:mtcurves}. Our AG model converges to the same loss as the BASE model, but the Hard-coded Gaussian model~\cite{you2020hard} converges to a slightly higher loss.

\subsection{Model Configurations}

We follow~\citet{liu2019roberta} for all design choices not mentioned in Table~\ref{appendix:table:model}. The size of feed-forward layers is always $4\times d_{model}$.

\begin{table}[H]
\centering
\resizebox{\columnwidth}{!}{
\begin{tabular}{l c c c}
\toprule
\textbf{Model} & \textbf{Layers} & \textbf{Heads} & \textbf{Hidden Size} \\ \midrule 

RoBERTa-8 & 8 & 12 & 768 \\
RoBERTa-12 & 12 & 12 & 768 \\
RoBERTa-16 & 16 & 16 & 768 \\
\bottomrule
\end{tabular}
}
\caption{Model design choices. Hidden Size is $d_{model}$ in~\citet{vaswani2017attention}. Heads is the number of heads per layer.}
\label{appendix:table:model}
\end{table}

\subsection{Best performing hyperparameters} \label{appendix:hyperparameters}

The best performing hyperparameters for each model are mentioned in Table~\ref{appendix:table:best}. All design choices that are not mentioned (like dropout in the feed-forward layer) follow \citet{liu2019roberta}.

\begin{table*}[t]
\centering
% \resizebox{\linewidth}{!}{
\begin{tabular}{clccccc}\toprule
\textbf{Language} & \textbf{Model} & \textbf{Learning Rate} & \textbf{Warmup Steps} & $\mathbf{\alpha_0/\lambda}$ & \textbf{Batch Size} \\ \midrule
\multirow{6}{*}{\textbf{English}} & RoBERTa-8-MLM  & 1e-4 & 10000 & - & 120 \\
 & RoBERTa-12-MLM  & 5e-5 & 10000 & - & 84 \\
 & RoBERTa-16-MLM  & 1e-5 & 10000 & -  & 48\\ \cdashlinelr{2-6}
 & RoBERTa-8-AG & 1e-4 & 0 & 100/0.5 & 120 \\
 & RoBERTa-12-AG  & 1e-4 & 0 & 10/0.5 & 84 \\
 & RoBERTa-16-AG & 1e-4 & 0 & 10/0.5 & 48 \\
 \midrule
\multirow{6}{*}{\textbf{Filipino}} & RoBERTa-8-MLM  & 1e-4 & 10000 & - & 40 \\
 & RoBERTa-12-MLM  & 5e-5 & 10000 & - & 28 \\
 & RoBERTa-16-MLM  & 1e-5 & 10000 & - & 16 \\ \cdashlinelr{2-6}
 & RoBERTa-8-AG & 1e-4 & 0 & 100/0.5 & 40 \\
 & RoBERTa-12-AG  & 1e-4 & 0 & 10/0.5 & 28
 \\
 & RoBERTa-16-AG & 1e-4 & 0 & 10/0.5 & 16 \\
\midrule
\multirow{6}{*}{\textbf{Oromo}} & RoBERTa-8-MLM  & 1e-4 & 1000 & - & 40 \\
 & RoBERTa-12-MLM  & 5e-5 & 1000 & - & 40 \\
 & RoBERTa-16-MLM  & 1e-5 & 1000 & - & 32 \\ \cdashlinelr{2-6}
 & RoBERTa-8-AG & 1e-4 & 0 & 100/0.5 & 40 \\
 & RoBERTa-12-AG  & 1e-4 & 0 & 10/0.5 & 40 \\
 & RoBERTa-16-AG & 1e-4 & 0 & 10/0.5 & 32 \\
\bottomrule
\end{tabular}
% }
\caption{Best performing hyperparameters. $\alpha_0$ is the relative weight placed on AG loss (equation~\ref{equation:alpha}) and $\lambda$ is the fraction of heads being guided in each layer.}
\label{appendix:table:best}
\end{table*}

\end{document}